\documentclass[]{IEEEtran}

\usepackage{microtype}
\usepackage{graphicx}
\usepackage{subfigure}
\usepackage{multirow} 
\usepackage{booktabs} 
\usepackage{algorithm}
\usepackage{algorithmic}
\usepackage{xspace}

\usepackage{hyperref}

\usepackage{amsmath}
\usepackage{amssymb}
\usepackage{mathtools}
\usepackage{amsthm}
\usepackage{stmaryrd}
\usepackage{mathtools}
\usepackage{colortbl}

\usepackage{hyperref}
\usepackage{url}
\usepackage{multirow}
\usepackage[flushleft]{threeparttable}
\usepackage{longtable}
\usepackage{supertabular}
\usepackage{colortbl}
\usepackage{xspace}
\usepackage{wrapfig}
\usepackage{xcolor}
\usepackage{cancel}
\usepackage{tikz}

\theoremstyle{plain}

\theoremstyle{definition}

\theoremstyle{remark}

\usepackage[textsize=tiny]{todonotes}

\newcommand{\bl}{\color{black}}
\newcommand{\modelname}{KARL\xspace}

\title{Improving Molecule Generation and Drug Discovery with a Knowledge-enhanced Generative Model}
\author{Aditya Malusare and Vaneet Aggarwal\thanks{The authors are with the Edwardson School of Industrial Engineering and the Institute of Cancer Research, Purdue University, email: \{malusare, vaneet\}@purdue.edu. 

A. Malusare gratefully acknowledges the Walther Cancer Foundation and support from the Purdue University Institute for Cancer Research, P30CA023168.

This work uses the Anvil supercomputer at Purdue University through allocation CIS230228 from the Advanced Cyberinfrastructure Coordination Ecosystem: Services \& Support (ACCESS) program. 

This paper has been accepted for publication at the IEEE/ACM Transactions on Computational Biology and Bioinformatics (October 2024). 

© 2024 IEEE. Personal use of this material is permitted. Permission from IEEE must be obtained for all other uses, in any current or future media, including reprinting/republishing this material for advertising or promotional purposes, creating new collective works, for resale or redistribution to servers or lists, or reuse of any copyrighted component of this work in other works.

}
}

\begin{document}
\maketitle

\begin{abstract} 

Recent advancements  in generative  models have established state-of-the-art benchmarks in the generation of molecules and novel drug candidates. Despite these successes, a significant gap persists between generative models and the utilization of extensive biomedical knowledge, often systematized within knowledge graphs, whose potential to inform and enhance generative processes has not been realized.  In this paper, we present a novel approach that bridges this divide by developing a framework for knowledge-enhanced generative models called \modelname. We develop a scalable methodology to extend the functionality of knowledge graphs while preserving semantic integrity, and incorporate this contextual information into a generative framework to guide a diffusion-based model. The integration of knowledge graph embeddings with our generative model furnishes a robust mechanism for producing novel drug candidates possessing specific characteristics while ensuring validity and synthesizability. \modelname outperforms state-of-the-art generative models on both unconditional and targeted generation tasks.

\end{abstract}


\begin{figure*}
	\includegraphics[width=\textwidth]{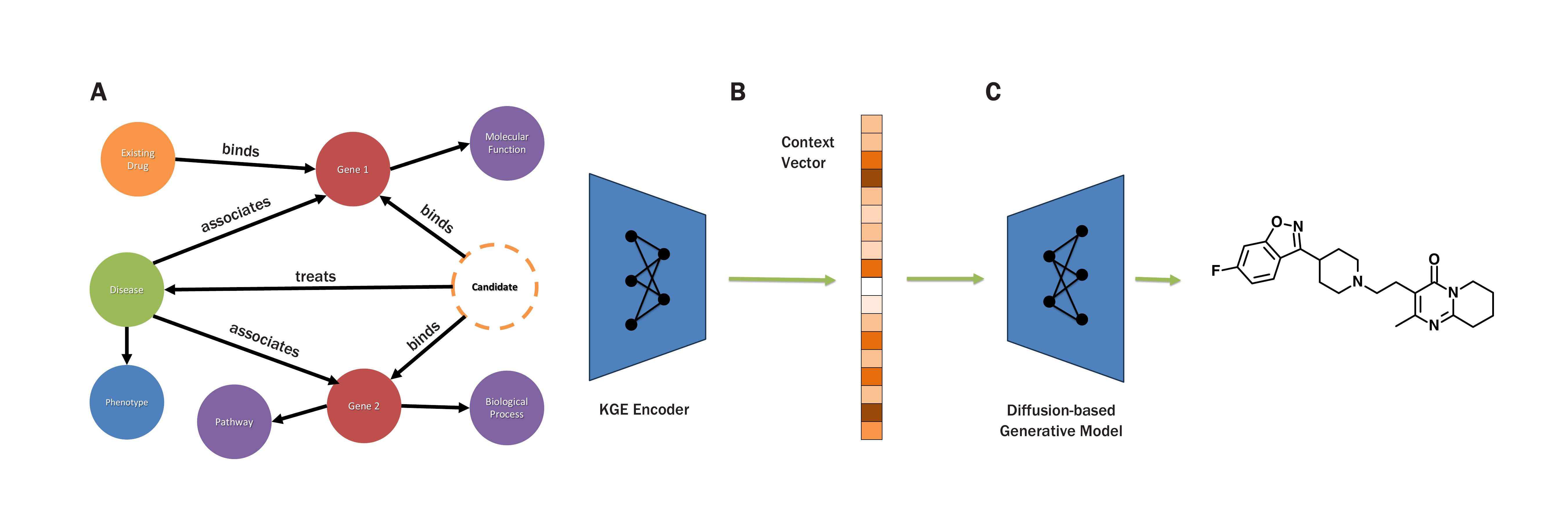}
    
    \caption{\textbf{\modelname Model} \textbf{A} Biomedical knowledge graph showing the links between drugs, target proteins, diseases, genes, biological pathways, etc. The edges represent different types of relationships like targeting, synergistic or unwanted interactions, or side effects. \textbf{B} We obtain Knowledge Graph Embeddings (KGEs) (Sec \ref{sec:kge}) for the target drug candidate. \textbf{C} These embeddings are used to guide a generative process (Sec \ref{sec:implement}).  \textbf{D} The end result of the generative process is a novel drug candidate. }
    \label{fig:main}
\end{figure*}

\section{Introduction}

Drug discovery is an expensive endeavor, with costs often surging beyond the billion-dollar mark, primarily due to the extensive stages involved in the development pipeline, ranging from target identification to clinical evaluations \cite{10.1016/j.healthpol.2010.12.002}. Nevertheless, the advent of computational techniques and machine learning has ushered in a paradigm shift in pharmaceutical research, streamlining drug development, mitigating expenses, and enhancing the discovery of bioactive compounds \cite{10.22270/jddt.v10i4.4218}. Predominantly, the emergence of sophisticated machine learning applications has substantially bolstered predictive capabilities, accelerating the pace of \textit{in-silico} drug design \cite{10.1093/bib/bby061}. However, this rapid progression has not been devoid of challenges. Present-day generative models in machine learning showcase impressive results but often miss out on tapping into the extensive biological knowledge available in the domain \cite{10.1021/acs.jcim.3c00465}. These models frequently grapple with issues like overfitting, lack of generalizability to new data \cite{10.1021/ci500747n}, and difficulties in handling novel drug or target interactions \cite{10.1093/bib/bbac269}. Although there is a vast amount of biomedical knowledge, there is a missing link between these collective datasets and the efforts to build generative models for drug discovery. 

Addressing these limitations, we introduce an innovative methodology for generating drugs that not only harnesses structured biological knowledge but ensures semantic coherence when generating new insights from existing knowledge graphs. Our scalable model employs knowledge graph embeddings to direct a diffusion-based generative model, fine-tuned by an original reinforcement learning reward strategy. This symbiosis between comprehensive bio-knowledge and generative models paves a pathway for the interpretative, scalable, and controllable generation of drug molecules. 

In order to improve the process of computational drug discovery, we focus on two key aspects: (i) \textbf{Improved generative models}: These models are capable of learning complex distributions of molecular structures from vast datasets and generating new compounds with desired properties. By automating the molecular design process, generative models can quickly propose viable drug candidates that match specific efficacy and safety profiles, potentially leading to breakthroughs in the treatment of diseases, and (ii) \textbf{Knowledge Graphs (KGs) of biomedical data: } KGs in drug discovery constitute a structured representation of vast amounts of heterogeneous data, encompassing the relations between different entities such as genes, proteins, drugs, and diseases. By leveraging knowledge graphs, we can create interpretable and generalizable models that encapsulate both the biomedical knowledge encoded within the relationships and the data-driven insights garnered from machine learning models. These graphs also enable a deeper analysis of biological pathways, off-target effects, and drug repurposing opportunities. 

Molecular generation is a critical aspect of drug discovery, material science, and chemical exploration. Generative models have demonstrated the ability to generate novel structures across a variety of representations, from 1-D SMILES strings to 2-D and 3-D molecular graphs. However, while these models have shown great results in producing drug-like molecules, they are typically evaluated based on their unconditioned generation capabilities. Generative models have also achieved significant improvements in targeted generation with modalities like text, images \cite{black2023training}, sound, and videos. The development of a targeted generative model for graphs is of particular interest in the problem of drug discovery. 

Biomedical knowledge can be organized and represented by a powerful tool called a Knowledge Graph. A good way of guiding the generative process is by using the KG as a structural and semantic scaffold, which can provide context-relevant constraints and objectives for the generative models. This integration of KGs with generative models can facilitate the targeted generation of molecules by driving the model to consider biological relevance and plausibility in the generation process. By doing so, the models can leverage both the relational knowledge embedded within the KG, such as the interactions between proteins, genes, drugs, and diseases, and the generative capabilities to propose novel compounds that are more likely to exhibit desired therapeutic effects and pharmacological profiles.

Incorporating KGs into generative modeling allows for a more informed exploration of the chemical space, focusing on areas with higher potential for successful drug development. The properties and relationships within KGs can guide the generative model in synthesizing molecules that not only are structurally novel but also align with known biological pathways and mechanisms of action. This coupling can lead to the development of generative models that produce candidates which are not only chemically valid but also biologically relevant. Moreover, such directed generation can help in hypothesis generation for drug repurposing and identifying previously unrecognized drug-target interactions, greatly enriching the pipeline of drug discovery and potentially reducing the timeline for the development of new therapies.


{ We propose an novel model capable of end-to-end generation (Figure \ref{fig:main}) called the \textbf{K}nowledge-\textbf{A}ugmented \textbf{R}einforcement \textbf{L}earning-optimized (\modelname) Model. This paper presents the following key innovations:  }

\begin{enumerate}
	\item A diffusion-based generative model for graphs steered with Discrete Diffusion Policy Optimization (DDPO) for improved molecule generation. The generative model, initially trained on extensive chemical structure datasets, is fine-tuned with the DDPO approach that rewards the model for producing chemically valid, pharmacologically active structures that have the ability to satisfy user-defined properties. (Section \ref{sec:molgenmol})
	\item A Knowledge Graph Embedding (KGE) model that generates contextual embeddings from large-scale biomedical databases that are used for targeted extrapolation and synthesis of drug candidates. By reformulating models that generate KGEs from energy-based models to ones that minimize a Maximum Likelihood Estimation (MLE) objective, we demonstrate the ability to generate more meaningful embeddings that also incorporate domain constraints, which ensure that predictions follow user-specified logical formulae. (Section \ref{sec:kge})
	\item The development of a training workflow that is used by the model to learn conditional generation, with the help of a property inference network that integrates the knowledge represented by the KGEs into the training of the diffusion model to create a unified process for knowledge-enhanced novel drug discovery (Section \ref{sec:implement})
\end{enumerate}

The novelty of our framework lies in: 1) the extension of the DDPO procedure from the fixed topology of images to the variable message-passing framework of graphs, 2) the creation of a KGE model with domain constraints specific to drug-based knowledge, and 3) the development of a mapping from molecule-space to KGE-space that is used for score-based conditional generation. 

\section{Related Work}

\textbf{Molecular Generation. } 

Molecular representations can be classified into three broad categories: (i) string-based (1-D), (ii) graph-based (2-D), and coordinate-based (3-D) representations. The simplified molecular-input line-entry system (SMILES) \cite{weininger1988smiles} remains one of the most widespread implementations of the 1-D representation. SMILES has been used with a variety of models such as Variational Auto-encoders (VAEs) \cite{jin2018junction} \cite{gomez2018automatic}, Generative Adversarial Networks (GANs) like \cite{tang2023earlgan}, and Recurrent Neural Networks (RNNs) \cite{neil2018exploring}. The simplified nature of SMILES strings does not explicitly capture geometrical and topological features. Additionally, small changes in the strings can lead to vastly different or invalid molecular structures \cite{lavecchia2019deep}, which is of considerable concern while using diffusion-based models. In order to mitigate these problems, alternative models like \cite{yuksel2023selformer} \cite{danel2023docking} using the SELF-referencIng Embedded Strings (SELFIES) \cite{krenn2020self} have been developed which ensure validity and complete coverage of chemical space.

Two-dimensional representations present the opportunity of directly working with the structure of compounds. Although they do not take into account the 3-D structure of the molecule, 2-D graph-based representations have achieved impressive results using models like {\bl GANs \cite{martinkus2022spectre}}, normalizing flows \cite{shi2019graphaf} \cite{zang2020moflow}, and diffusion-based models \cite{jo2022score} \cite{vignac2023digress} \cite{kong2023autoregressive} \cite{guo2023diffusing}. 

\textbf{Knowledge-enhanced Models. } 

Knowledge graphs have gained traction in drug discovery \cite{James2023} by structuring diverse biological data into interconnected frameworks that facilitate the identification of new therapeutic targets and drug repurposing opportunities \cite{Santos2022}, \cite{Zheng2020}, \cite{Chandak2023}. These graphs represent entities such as molecules, proteins, and diseases as nodes, with edges capturing the connections and associations between them by encoding predicates like {\bl `causes,' `indicates,' `interacts,' `targets,'} etc. The paper by \cite{yu2021predicting} showcases the use of entity embeddings from knowledge graphs for predicting drug-disease associations, while \cite{zitnik2018modeling} employed graph neural networks on similar structures to uncover drug-target interactions, surpassing traditional predictive models. By incorporating relational information from knowledge graphs, molecular generative models can impose biologically relevant constraints during the generation process \cite{bilodeau2022generative}. Knowledge graphs also aid in systematically identifying adverse drug reactions \cite{novavcek2020predicting} and repurposing existing drugs by analyzing the network's topology to reveal hidden biological pathways \cite{pan2022deep}.

\section{Molecular Generative Model}
\label{sec:molgenmol}

\begin{figure*}
    \centering
    \includegraphics[width=0.95\textwidth]{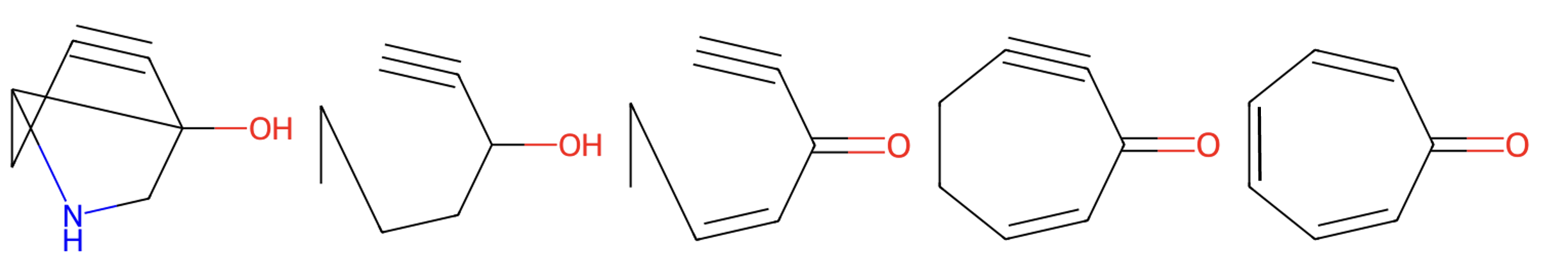}
    \caption{\textbf{Generative process. } The random initialization of the molecular graph \textbf{G} converges to a valid non-benzeneoid aromatic compound called tropone through the diffusion model.}
    \label{fig:reverseprocess}
\end{figure*}

Molecular structures can be represented using a planar graph $\textbf{G} = (\textbf{X}, \textbf{A})$ where $\textbf{X}\in\mathbb{R}^{N\times M}$ is a feature matrix for $N$ nodes (heavy atoms) described by $M$-dimensional vectors encoding atom information, and $\textbf{A}\in\mathbb{R}^{N\times N}$ is the adjacency matrix indicating the presence of single, double or triple bonds between the nodes. 

Our generative model is built on the foundations of the GDSS \cite{jo2022score} and MOOD \cite{Lee2022ExploringCS} diffusion models. Graph Diffusion via the System of SDEs (GDSS) defines the forward diffusion $q$ of a graph $ \textbf{G}_t = (\textbf{X}_t, \textbf{A}_t) $ with a  Stochastic Differential Equation (SDE):

\begin{equation}
\label{eq:sde}
	\mathrm{d}\textbf{G}_t = \textbf{f}_t(\textbf{G}_t) + g_t\mathrm{d}\textbf{w}
\end{equation}

where $\textbf{w}$ is the standard Wiener process and $\textbf{f}_t$ and $g_t$ are the coefficients of linear drift and scalar diffusion, respectively. GDSS performed well at generating molecular graphs, but its distributional learning led to generated molecules closely resembling the training dataset. The Molecular Out-of-distribution (MOOD) framework overcame the restricted exploration space of the training process by modifying the above equation into a conditional SDE, with the marginal distribution of the forward process becoming $p_{\theta}(\textbf{G}_t|\textbf{y}_o=\lambda)$. Here, $\textbf{y}_o$ represents the OOD condition and the hyperparameter $\lambda\in[0,1)$ tunes the ``OOD-ness" of the generative process. Controlling $\lambda$ allows MOOD to explore areas outside the training space and generate novel molecules.  

Since we aim to use contextual information from KGs for our generative model, we formulate an extension that samples from a more general marginal distribution $p_{\theta}(\textbf{G}_t|\textbf{c})$ conditioned on contextual data $\textbf{c}$. We then introduce a RL-based objective to maximize a reward $r$: 
\begin{equation}
	\max \mathbb{E}_{\textbf{c}\sim p(\textbf{c}), \textbf{G}_0 \sim p(\textbf{G}_0|\textbf{c})}\left[r(\textbf{G}_0, \textbf{c})\right] 
\end{equation}

Black et al. \cite{black2023training} introduce a fine-tuning technique known as Denoising Diffusion Policy Optimization (DDPO) for image-based models. This method excels in handling images by incorporating human feedback to ensure metrics like aesthetic qualities—a factor typically difficult to quantify in computational models. Despite its success in the domain of images, extending these techniques to the realm of molecule optimization presents significant challenges. The intricate dependencies between nodes and edges in molecular structures demand careful consideration, as these relationships are critical to determining the validity and inherent properties of the molecules. 

We introduce a reward function $r$ that simultaneously prioritizes multiple molecular properties: (i) The drug-likeness $Q$ of a molecule, quantified by its QED score \cite{bickerton2012quantifying}, (ii) the synthesizability $S$ calculated using the SAScore \cite{ertl2009estimation} which is a rule-based determination of synthetic accessibility, and (iii) a provision for defining a property $C$ in the chemical space like novelty/molecular similarity, docking scores, structure validity, etc. 

We define our overall reward function as: 

\begin{equation}
\label{eq:rewardfunc}
	r(\textbf{G}, \textbf{c}) = 
\begin{cases}

   \kappa_1 Q(\textbf{G}) + \kappa_2 S(\textbf{G}) + \kappa_3 C(\textbf{G}, \textbf{c}); & \\  
    -\kappa_4, \hspace{30pt} \text{if \textbf{G} is invalid} & 
\end{cases}
\end{equation}

We elaborate on the training process of the model for conditional generation in Section \ref{sec:implement}. 

\section{Knowledge Graph Embeddings}
\label{sec:kge}

A knowledge graph $\mathcal{G}$, involving a set of entities $\mathcal{E}$ and relations $\mathcal{R}$, is a directed multigraph composed of triples $(s, r, o) \in \mathcal{E}\times\mathcal{R}\times\mathcal{E}$ that represent the relation $r$ between subject $s$ and object $o$ nodes. In biomedical knowledge graphs, $\mathcal{E}$ contains entities like drugs, diseases, genes, phenotypes, and biological pathways while $\mathcal{R}$ contains relations like interactions, drug targets and side effects. Some examples of public domain knowledge graphs in this field like Hetionet \cite{himmelstein2017systematic}, CKG \cite{Santos2022}, BioKG \cite{walsh2020biokg}, PharmKG \cite{Zheng2020}, and PrimeKG \cite{Chandak2023}, which range from $\sim$10K to 10M nodes and up to $\sim$200M relations. 

Knowledge Graph Embeddings (KGEs) are low dimensional representations of relations and entities that are used for various tasks like link prediction, graph completion and attribute inference. KGEs are constructed with a scoring function $\phi_{\mathcal{G}}: \mathcal{E}\times\mathcal{R}\times\mathcal{E} \to \mathbb{R}$ that measures the plausibility of any given triple. 

KGEs help preserve the structure and information of neighbors, enabling the efficient encoding of the topology and semantic relationships inherent in knowledge graphs. Particularly in the biomedical domain, where the complexity of interactions is high, the capacity of KGEs to facilitate the extraction of latent relationships between disparate entities is crucial for advancements in drug repurposing, disease gene prioritization, and patient outcome prediction.

KGE models involve various translational distance-based metrics like TransE \cite{lin2015learning}, tensor factorization methods such as RESCAL \cite{nickel2011three}, and neural network-based models like ConvE \cite{dettmers2018convolutional}. These models differ in how they interpret the relationships and interactions between entities, and each comes with relative strengths and weaknesses with regard to specific types of relational data and structures. 

Traditionally, score-based KGE models have been interpreted as energy-based models, where the score is seen as a measure of the negative energy of an $(s, r, o)$ triplet. Recasting these models into a probabilistic interpretation would enable exact training by Maximum Likelihood Estimation (MLE), as well as the ability to encode domain constraints into the learning process.  In order to transform these negative energies into probabilities over the $\mathcal{E}\times\mathcal{R}\times\mathcal{E}$ space, we need to calculate the partition function, which is infeasible for large-scale biomedical KGs. \cite{loconte2023how} demonstrate that KGE models can be represented by structured computational graphs, called circuits, which are expressive probabilistic functions over the triplet space and can be efficiently trained with MLE. This enables us to use KGE models as efficient generative models of new $(s, r, o)$ triples, consistent with the statistics of existing the KG while guaranteeing the satisfaction of constraints that would be crucial to  pharmacological applications.  

The MLE objective is evaluated as:

\begin{equation}
  \begin{aligned}
	\mathcal{L}_{MLE} & = \sum_{(s, r, o) \in \mathcal{G}} \log{p(s, r, o)} \\ 
	       & = -|\mathcal{G}|\log{Z} + \sum_{(s, r, o) \in \mathcal{G}} \phi_{pc}(s, r, o)  
	     \end{aligned}
\end{equation}

Circuit-based score functions  bring down the complexity of evaluating $Z$ from $\mathcal{O}(|\mathcal{E}|^2 \cdot |\mathcal{R}|\cdot \text{cost}(\phi))$ to $\mathcal{O}(|\mathcal{E} + \mathcal{R}|\cdot \text{cost}(\phi))$ \cite{loconte2023how}. 

The generation of new $(s, r, o)$ triplets is ensured to be semantically coherent within the rules of the KG by introducing domain constraints \cite{ahmed2022semantic} that modify the score functions with indicator functions for valid triples. Given a relation $r\in\mathcal{R}$ the subsets $S_r, O_r\subset\mathcal{E}$ represent the sets of entities between which $r$ is a valid relation. This domain 	$K_r = S_r \wedge \{r\} \wedge O_r$ can be extended across all $r$ to form $K=\vee_r K_r$. We restrict the score function to only valid triples $(s, r, o)\in K$ by introducing an indicator function $c_K$ in the score calculation $\phi(s,r,o) = p(s,r,o)\cdot c_K(s,r,o)$. For example, in our experiments, we are particularly interested in drug-protein interaction. For this type of relation $r=\texttt{targets}$, we restrict $s\in\texttt{drugs}$ and $o\in\texttt{proteins}$. Once again, the circuit-based score functions help us train these models with a complexity of $\mathcal{O}(|\mathcal{E} + \mathcal{R}|\cdot \text{cost}(\phi) \cdot \text{cost}(c_K) )$.

In order to generate KGEs (Fig \ref{fig:main} C), \modelname uses the RotatE \cite{sun2019rotate} algorithm with the above modifications to the training process in order to restrict the model to generate coherent triples. The embeddings generated from the knowledge graph are used to guide the generative process as described below.

\section{Implementation of \modelname}
\label{sec:implement}

An essential aspect of drug design is the ability to specify the target properties in our generative process. This section describes our implementation of the guidance scheme for our diffusion model, which we refer to as the conditional diffusion model. We first describe the creation of a regressor that predicts properties based on graph structure, and then formulate the conditional process that guides the diffusion model to push it to generate molecules with the desired properties. 

\subsection{Property Inference Network $P_{\phi}(\textbf{G})$}

To guide the conditional generation process, we create a neural network to estimate knowledge-based embeddings $\textbf{c}$ from a noised version of an input molecular graph $\textbf{G}_T$. $P_{\phi}(\textbf{G}_T)\approx\textbf{c}$ is used to implement a modified version of the classifier guidance algorithm by \cite{sohl2015deep}. While previous work by \cite{Lee2022ExploringCS} and \cite{vignac2023digress} uses a similar algorithm to guide conditional generative processes, ours is a novel approach that utilizes a combination of graph attention and convolutional layers to estimate knowledge-based  embeddings, effectively creating a map between chemical space and KGE space.

 $P{_\phi}(\textbf{G}) = P_{\phi}(\textbf{X}, \textbf{A})$ is constructed by first passing the feature $\textbf{X}$ and adjacency matrices $\textbf{A}$ through an aggregation operation: 
 
 \begin{equation}
 	H^1 = \sigma(\textbf{A}\textbf{X}W^0_{\phi})
 \end{equation}
 
 We then use a stack of self-attention layers:
 \begin{equation}
 	h_i^{l+1} = \sum_{j}\alpha_{ij} ~\cdot~ \mathbf{W^l}h_j^l
 \end{equation}
 where
 \begin{equation}
 	\alpha_{ij} = \text{SoftMax}(\sigma(\mathbf{W_a}^l \cdot [\mathbf{W}^l h^l_i ~\oplus~ \mathbf{W}^l h^l_j]))
 \end{equation}

Here, $\textbf{W}^l, \textbf{W}_a^l$ are learnable parameters at the $l$-th layer. The stack of attention layers produces a final output of dimension $|\textbf{c}|$. (Training details in Appendix \ref{appendix:training})

\subsection{Conditional Diffusion Training}

The stochastic forward process described in Eq. \eqref{eq:sde} can be used for generation by solving it's reverse-time version (Fig \ref{fig:reverseprocess}):

\begin{equation}
\label{eq:reversesde}
	\mathrm{d}\textbf{G}_t = [\textbf{f}_t(\textbf{G}_t)-g_t^2 \nabla_{\textbf{G}_t} \log{p_{\theta}(\textbf{G}_t})]\mathrm{d}\overline{t} + g_t\mathrm{d} \overline{\textbf{w}}
\end{equation}

where $\overline{t}$ and $\overline{\textbf{w}}$ represent a reverse time-step and stochastic process. A score network $s_{\theta}$ is used to approximate $\nabla_{\textbf{G}_t} \log{p_t(\textbf{G}_t})$ and simulate the reverse process  in time, to generate $\textbf{G}_{t-1}$. 

Conditioning this process can be achieved by adding the conditioning information $\textbf{c}$ at each diffusion step: $p_{\theta}(\textbf{G}_t|\textbf{c})$. The score network is then used to approximate the modified gradient $\nabla_{\textbf{G}_t} \log{p_{\theta}(\textbf{G}_t}|\textbf{c})$. The conditional distribution is rearranged to give 

\begin{equation}
	\nabla_{\textbf{G}_t} \log{p_{\theta}(\textbf{G}_t}|\textbf{c}) \propto \nabla_{\textbf{G}_t}\log p_{\theta}(\textbf{G}_t) + \nabla_{\textbf{G}_t}\log p_{\theta}(\textbf{c}|\textbf{G}_t)
\end{equation}

The term $\nabla_{\textbf{G}_t}\log p_{\theta}(\textbf{c}|\textbf{G}_t)$ in the above equation steers the model towards optimizing for the condition, while the first term $\nabla_{\textbf{G}_t}\log p_{\theta}(\textbf{G}_t)$ introduces variation into the trajectory and helps explore newer regions. 

We model the conditional probability using a distribution of the form:
\begin{equation}
	p_{\theta}(\textbf{c}|\textbf{G}) = \dfrac{1}{Z_{\theta}}\exp({-\alpha_{\theta}||\textbf{c} - P_{\phi}(\textbf{G})||^2})
\end{equation}

where $\alpha_{\theta}$ is a scaling factor and $Z_{\theta}$ is the partition function. This procedure has a tractable complexity, supported by the analysis in Section \ref{sec:kge}. 

\section{Experiments}

We perform individual evaluations of each component, followed by an end-to-end study of \modelname in synthesizing novel drug candidates. We compare unconditional generation against variants of generative models that use GANs \cite{martinkus2022spectre}, diffusion \cite{jo2022score, vignac2023digress}, autoregressive methods \cite{kong2023autoregressive}, and normalizing flows \cite{shi2019graphaf, zang2020moflow}.

\subsection{Evaluating Unconditional Molecular Generation}

\textbf{Experimental Setup.  } The Quantum Machines 9 (QM9) \cite{ramakrishnan2014quantum} and ZINC \cite{irwin2012zinc} datasets contain 134k and 250k chemical structures and are designed to aid in the exploration of chemical space. We follow previous work \cite{kong2023autoregressive} and perform unconditional generation of 10,000 molecular structures. We report the percentage of valid and unique molecules, along with the novelty as defined by \cite{jin2020multi} and the Frechet ChemNet Distance (FCD) \cite{preuer2018frechet}, which evaluates the similarity between the generated and training datasets using the activations of the penultimate layer of the ChemNet. The reward function (Eq. \eqref{eq:rewardfunc}) is used to fine-tune \modelname for maximizing the validity and novelty of the molecules. 

\textbf{Results.  } Tables \ref{table:qm9} and \ref{table:zinc} demonstrate that \modelname achieves state-of-the-art results on all metrics except the FCD on the QM9 dataset. We omit reporting uniqueness on ZINC since all models were $\gtrapprox 99.99\%$ unique. Previous results sourced from \cite{kong2023autoregressive}. \modelname generates valid molecules at rates of 99.25\% and 98.29\% on QM9 and ZINC, respectively as a result of the fine-tuning procedure. 
 
\begin{table}[!t]
    \centering
    \footnotesize
    \setlength{\tabcolsep}{0.3em}
    \begin{tabular}{@{}lccccc@{}}
    \toprule
    \textbf{Model}   & Validity$\uparrow$  & FCD$\downarrow$    & Unique$\uparrow$ & Novelty$\uparrow$  \\ 
    \midrule
    GraphAF  & 74.43      & 5.27  & 88.64      & 86.59   \\
    MoFlow   & 91.36      & 4.47  & 98.65      & 94.72   \\
    SPECTRE  & 87.3       & 47.96 & 35.7       & 97.28   \\
    GDSS     & \underline{95.72}      & 2.9    & 98.46      & 86.27 \\  
    DiGress  & \textbf{99.0}      & \textbf{0.36}  & 96.66      & 33.4   \\ 
    GraphARM& 90.25     & \underline{1.22}   & 95.62      & 70.39  \\ 
    \midrule
    \modelname (w/o RL)& 94.20    & 2.21   & 97.64      & 95.63  \\ 
    \modelname& \textbf{99.25}     & 1.54   & 98.62      & \textbf{99.74}  \\ 
    \bottomrule
    \end{tabular}
    \vspace{4pt}
    \caption{\textbf{Generation results on the QM9 dataset.} (\textbf{Best}, \underline{Second})}
    
    \label{table:qm9}
 \end{table}

\begin{table}[!t]
    \centering
    \footnotesize
    \setlength{\tabcolsep}{0.3em}
    \begin{tabular}{@{}lccccc@{}}
    \toprule
    \textbf{Model}   & Validity$\uparrow$  & FCD$\downarrow$    & Unique$\uparrow$   \\ 
    \midrule
    GraphAF  & 68.47      & \underline{16.02}  & 98.64     \\
    MoFlow   & 63.11      & 20.93  & 99.99     \\
    SPECTRE  & 90.2       & 18.44 & 67.05     \\
    GDSS     & \underline{97.01}      & \textbf{14.66}    & 99.64  \\  
    DiGress  & 91.02      & 23.06  & 81.23     \\ 
    GraphARM& 88.23     & \underline{16.26}   & 99.46    \\ 
    \midrule
    \modelname (w/o RL)& 96.17    & 17.77  & 97.64        \\ 
    \modelname& \textbf{98.29}     & \textbf{14.56}       & 99.95  \\ 
    \bottomrule
    \end{tabular}
    \vspace{4pt}
    \caption{\textbf{Generation results on ZINC.} (\textbf{Best}, \underline{Second})}
    
    \label{table:zinc}
    \end{table}

\subsection{Drug-based Knowledge Graph Embeddings}

\textbf{Experimental Setup.  } \cite{Bonner2022} devise a procedure to assess how well biomedical KGE models can predict missing links in knowledge graphs, standard evaluation employs knowledge graph metrics like Mean Reciprocal Rank and Hits@k. These metrics measure the model's ability to rank true triples higher than corrupted ones in link prediction tasks.  Models are  evaluated on predicting both head and tail entities by corrupting each in turn. Robustness is examined by testing across multiple random parameter initializations and dataset splits. Performance is evaluated on real-world KGs - BioKG \cite{walsh2020biokg} and Hetionet \cite{himmelstein2017systematic}. The following metrics are used for comparison: (i) Mean Reciprocal Rank (MRR) (ii) Hits@1 and Hits@10 (iii) Adjusted Mean Rank \cite{berrendorf2020ambiguity}. Hits@k is computed using the following procedure: For each corrupted triple (with head or tail entity removed) in the test set, the model ranks all possible candidate entities to complete the triple. A ``hit" is when the true missing entity is ranked within the top $k$ predictions. Hits@k calculates the proportion of test triples where the true entity was ranked in the top $k$.

\textbf{Results.  } Table \ref{tab:model-seed} presents the mean performance over 10 random seeds of all models and both datasets as measured by the above metrics. Only the random seed used to initialize the model parameters is changed. The mean score of \modelname outperforms all metrics except the AMR score on BioKG, showing the improved quality of embeddings due to the introduction of domain constraints. We compare against ComplEx \cite{trouillon2017knowledge}, DistMult \cite{zhang2018knowledge}, RotatE \cite{sun2019rotate}, and TransE \cite{lin2015learning}. 

\begin{table*}[h!]
    \centering

    \begin{tabular}{l l  c c c  c c c  }
        \toprule
        \textbf{Dataset}          & \textbf{Approach} & \multicolumn{4}{c}{\textbf{Metric}}                                                                                         \\
        \midrule \midrule
                                  &                   & AMR \(\downarrow\)                       & MRR \(\uparrow\)           & Hits@1 \(\uparrow\)        & Hits@10 \(\uparrow\)     \\
        \cline{3-6}

        \multirow{6}{*}{Hetionet} & ComplEx           & 0.167\(\pm\)0.009                        & 0.026\(\pm\)0.009          & 0.008\(\pm\)0.003          & 0.064\(\pm\)0.024        \\
                                  & DistMult          & 0.201\(\pm\)0.303                        & 0.036\(\pm\)0.019          & 0.012\(\pm\)0.007          & 0.079\(\pm\)0.045          \\
                                  & RotatE            & 0.035\(\pm\)0.000               &  \textbf{0.127}\(\pm\)0.000					 & 0.063\(\pm\)0.000				 & 0.262\(\pm\)0.001 \\
                                  & TransE            & 0.053\(\pm\)0.000                        & 0.079\(\pm\)0.001          & 0.034\(\pm\)0.001          & 0.117\(\pm\)0.002          \\
                                  & TransH            & 0.126\(\pm\)0.000                        & 0.033\(\pm\)0.001          & 0.007\(\pm\)0.000          & 0.088\(\pm\)0.002        \\
                                  & \modelname  & \textbf{0.030}\(\pm\)0.002				& \textbf{0.125}\(\pm\)0.002			& \textbf{0.071}\(\pm\)0.000	& \textbf{0.33}\(\pm\)0.010	\\
        \midrule
        \multirow{6}{*}{BioKG}    & ComplEx           & 0.213\(\pm\)0.011                        & 0.008\(\pm\)0.001          & 0.003\(\pm\)0.000          & 0.008\(\pm\)0.003        \\
                                  & DistMult          & 0.560\(\pm\)0.339                        & 0.015\(\pm\)0.003          & 0.007\(\pm\)0.001          & 0.027\(\pm\)0.006          \\
                                  & RotatE            & \textbf{0.022}\(\pm\)0.000                        & 0.123\(\pm\)0.000          & 0.059\(\pm\)0.000 		   & 0.240\(\pm\)0.001 \\
                                  & TransE            & 0.021\(\pm\)0.000               & 0.062\(\pm\)0.000          & 0.019\(\pm\)0.000          & 0.134\(\pm\)0.001          \\
                                  & TransH            & 0.078\(\pm\)0.001                        & 0.022\(\pm\)0.000          & 0.008\(\pm\)0.000          & 0.042\(\pm\)0.001        \\
                                  & \modelname  & 0.025\(\pm\)0.010				& \textbf{0.136}\(\pm\)0.010			& \textbf{0.065}\(\pm\)0.000	& \textbf{0.28}\(\pm\)0.001	\\

        \bottomrule
    \end{tabular}
    \vspace{5pt}
    \caption{\textbf{Evaluation of KGE Model. } Previous results sourced from \cite{Bonner2022}.}
    \label{tab:model-seed}

\end{table*}

\begin{table*}[t!]
    \centering
    \resizebox{0.9\textwidth}{!}{
    \renewcommand{\arraystretch}{0.8}
    \renewcommand{\tabcolsep}{1.5mm}
    \begin{tabular}{l@{\hskip 0.1in}ccccc}
    \toprule
        \multirow{2.5}{*}{Method} & \multicolumn{5}{c}{Target protein} \\
    \cmidrule(l{2pt}r{2pt}){2-6}
        & parp1 & fa7 & 5ht1b & braf & jak2 \\
    \midrule
        REINVENT~\cite{olivecrona2017molecular} & \phantom{0}-8.702~\scriptsize($\pm$~0.523) & -7.205~\scriptsize($\pm$~0.264) & \phantom{0}-8.770~\scriptsize($\pm$~0.316) & \phantom{0}-8.392~\scriptsize($\pm$~0.400) & \phantom{0}-8.165~\scriptsize($\pm$~0.277) \\
        MORLD~\cite{jeon20morld} & \phantom{0}-7.532~\scriptsize($\pm$~0.260) & -6.263~\scriptsize($\pm$~0.165) & \phantom{0}-7.869~\scriptsize($\pm$~0.650) & \phantom{0}-8.040~\scriptsize($\pm$~0.337) & \phantom{0}-7.816~\scriptsize($\pm$~0.133) \\
        HierVAE~\cite{jin2020hierarchical} & \phantom{0}-9.487~\scriptsize($\pm$~0.278) & -6.812~\scriptsize($\pm$~0.274) & \phantom{0}-8.081~\scriptsize($\pm$~0.252) & \phantom{0}-8.978~\scriptsize($\pm$~0.525) & \phantom{0}-8.285~\scriptsize($\pm$~0.370) \\
        FREED~\cite{yang21freed} & -10.427~\scriptsize($\pm$~0.177) & \textbf{-8.297}~\scriptsize($\pm$~0.094) & -10.425~\scriptsize($\pm$~0.331) & -10.325~\scriptsize($\pm$~0.164) & \phantom{0}-9.624~\scriptsize($\pm$~0.102) \\
        FREED-QS & -10.579~\scriptsize($\pm$~0.104) & \textbf{-8.378}~\scriptsize($\pm$~0.044) & -10.714~\scriptsize($\pm$~0.183) & -10.561~\scriptsize($\pm$~0.080) & \phantom{0}-9.735~\scriptsize($\pm$~0.022) \\
        GDSS~\cite{jo22GDSS} & \phantom{0}-9.967~\scriptsize($\pm$~0.028) & -7.775~\scriptsize($\pm$~0.039) & \phantom{0}-9.459~\scriptsize($\pm$~0.101) & \phantom{0}-9.224~\scriptsize($\pm$~0.068) & \phantom{0}-8.926~\scriptsize($\pm$~0.089) \\
        MOOD  & \textbf{-10.865}~\scriptsize($\pm$~0.113) & -8.160~\scriptsize($\pm$~0.071) & \textbf{-11.145}~\scriptsize($\pm$~0.042) & \textbf{-11.063}~\scriptsize($\pm$~0.034) & \textbf{-10.147}~\scriptsize($\pm$~0.060) \\
        \midrule
        \textsc{\modelname} (w/o KGE conditioning)  & {-9.416}~\scriptsize($\pm$~0.463) & {-7.920}~\scriptsize($\pm$~0.321) & {-9.679}~\scriptsize($\pm$~0.345) & {-9.465}~\scriptsize($\pm$~0.098) & {-9.782}~\scriptsize($\pm$~0.303) \\
        \textsc{\modelname}  & \textbf{-11.475}~\scriptsize($\pm$~0.410) & \textbf{-8.270}~\scriptsize($\pm$~0.211) & \textbf{-11.419}~\scriptsize($\pm$~0.072) & \textbf{-11.364}~\scriptsize($\pm$~0.035) & \textbf{-10.636}~\scriptsize($\pm$~0.211) \\
    \bottomrule
    \end{tabular}}
    \vspace{4pt}
        \caption{ \textbf{Top 5\% docking score (kcal/mol). } The results are the means and the standard deviations of 5 runs. Previous results sourced from \cite{Lee2022ExploringCS}.}		
    \label{tab:novel_top5}
\end{table*}

\subsection{Knowledge-enhanced Drug Discovery: Novelty and Protein Targeting}

\textbf{Experimental Setup.  } \modelname generates 3,000 drug candidates with the reward function (Eq. \eqref{eq:rewardfunc}) calibrated to optimize synthesizability, drug-likeness and binding affinity to a target protein. {\bl The knowledge graph used to generate embeddings and train the Property Inference Network is BioKG \cite{walsh2020biokg}.}  Following \cite{Lee2022ExploringCS}, we choose the targets: \texttt{parp1}, \texttt{fa7}, \texttt{5ht1b}, \texttt{braf}, \texttt{jak2}. We set the chemical property $C$ to emphasize novelty, introducing a penalty for a similarity score $>0.4$ with the training dataset. The knowledge embeddings are generated by asking the model to complete the triple (\_\_\_\_\_, \texttt{targets}, \texttt{protein}) for all five targets, searching over the molecule space for the highest scoring embeddings. These embeddings are used  for conditional generation as desctribed in Section \ref{sec:implement}.

\textbf{Results.  } Table \ref{tab:novel_top5} reports the average Docking Score (DS) of the top 5\% of generated molecules. \modelname is able to generate novel molecules with the best docking scores among the models we compare against, with previous results sourced from \cite{Lee2022ExploringCS}. The procedure for calculating these scores is described in Appendix \ref{appendix:training}.

\subsection{Ablation Study: Generation without Reinforcement Learning}

We perform an ablation study to examine the effect of fine-tuning \modelname with the reward function (Eq. \eqref{eq:rewardfunc}) by comparing the performance before the fine-tuning procedure in Tables \ref{table:qm9} and \ref{table:zinc}. We find that even without DDPO, \modelname has a similar performance as other diffusion- and SDE-based systems. DDPO improves the validity and novelty to $\geq 99\%$, an improvement of 4-5\% points over the non fine-tuned performance. 

\subsection{Ablation Study: Generation without Knowledge Enhancement}

We validate the impact of integrating KGEs into the model by running the target protein experiments with a version of \modelname that is fine-tuned using DDPO but not trained for conditional generation. Table \ref{tab:novel_top5} shows that \modelname without KGE information obtains molecules that do not score as well in the docking metric. In most cases, the docking score is similar to that of GDSS, which is in line with our expectations, since our base model is derived from a similar framework. The DDPO fine-tuning is identical in both cases. We thus demonstrate that the incorporation of knowledge through embeddings leads to a measurable improvement in targeted drug parameters. We can also consider a comparison between \modelname and RotatE in Table \ref{tab:model-seed} as an ablation study, since our model is an improved version of RotatE with the modifications in Section \ref{sec:kge}.

\section{Conclusion}

We enhance molecular generation and drug discovery models by incorporating information from biomedical knowledge graphs into the generative process. This framework, called \modelname, demonstrates:
1) Performance on unconditional molecular generation on par with state-of-the-art generative models, 2) Improvements in Knowledge Graph Embedding (KGE) metrics by introducing domain constraints and reformulating the scoring process, and 3) Enhancement with KGEs improves targeted generation, evaluated with docking scores for five proteins.  

Through multiple experiments and ablation studies, we demonstrate that the extraction of embeddings from  knowledge graphs and their use in guiding generative models leads to a measurable improvement in the quality of generated drug candidates.

\bibliography{manual.bib, exported.bib}
\bibliographystyle{IEEEtran}


\newpage
\clearpage
\appendices

\setcounter{page}{1}
\section{Disclosure of Use of LLM Tools}

The authors use the claude.ai model from Anthropic to summarize papers and refine human-written text, and scite.ai to assist with literature review.  

\section{Implementation Details}
\label{appendix:training}

Our models are written in PyTorch 2.0, with the training workflow implemented using the Lightning framework. Hyperparameters are managed using Hydra. All experiments are performed on  machines with an Intel (R) Core i9-9980XE CPU and four NVIDIA (R) RTX 2080 Ti (12 GB) GPUs.  

\subsection{Knowledge graphs}

PyKEEN \cite{ali2021pykeen} (built using Pytorch) is used to implement the Knowledge Graph Embedding Model (KGEM). The knowledge graph dataloaders for BioKG and Hetionet in PyKEEN are used with a custom defined model based on the formulation of \cite{loconte2023how}. {\bl We implement TransE with an embedding dimension of 256. We use the scoring function $$f(h, r, t) = - \|\textbf{e}_h + \textbf{e}_r - \textbf{e}_t\|_{p}$$ with $p=1$. We use the Adam optimizer with a learning rate of 0.0001 and a batch size of 16384. The model is trained for 100 epochs. The model is trained using a Stochastic Local Closed World Assumption (SLCWA). }

\subsection{Docking scores}

We use AutoDock Vina to calculate docking scores, which reflect the predicted binding affinity between a receptor (protein) and a ligand (such as a drug molecule). We first prepare the receptor (protein) and ligand (drug) structure file in PDBQT format, which requires the removal of water molecules, addition of hydrogen atoms, assignment of charges, and determination of torsion angles. Once the PDBQT files are ready, the Python script defines the search space and creates a configuration file for Vina. The script then invokes the Vina command-line interface with the appropriate parameters, captures output, and saves the docking results, typically a PDBQT file containing the ranked binding poses according to their scores.

\subsection{Training the Property Inference Network}
\label{app:PIN}

The property inference network $P_\phi$ learns to estimate Knowledge Graph Embeddings (KGE), given a noised version of the graph, i.e., $P_{\phi}(\textbf{G}_t) \approx \textbf{c}$. We train this network using a learning schedule starting with the inputs being the original graphs $\textbf{G}_0$ and the targets set to their corresponding KGEs $\textbf{c}$. We then use the forward process $t$ times, where $t<=0.05\times S$, and $S$ is the number of generation steps. These noised graphs are similarly used as inputs with the label still being set to $\textbf{c}$ in order to train the network to approximate to this KGE in the neighborhood of the original graph $\textbf{G}_0$.

\subsection{Adapting Denoising Discrete Policy Optimization (DDPO) for Graphs}

We find that the DDPO formulation developed by Black et al. (2023) for images does not adapt well when directly applied to graphs, since the algorithm does not converge as the number of nodes increases. This can mainly be attributed to the variable topology of graphs and the presence of node and edge attributes that lead to an increase in fluctuations.

The MDP for our process is defined as follows:

$$ s_t = (\textbf{G}_t, T-t), a_t = \textbf{G}_{T-t-1} $$

$$ \pi_\theta(a_t|s_t) := p_\theta(\textbf{G}_{T-t-1} | \textbf{G}_{T-t}) $$

$$ r(s_t, a_t) = r(\textbf{G}_0) \text{ if } t = T; 0 \text{ otherwise} $$

The above framework is generally optimized using the following policy gradient:

$$ \nabla_{\theta}\mathcal{J}(\theta) = E_{\tau}\left[r(\textbf{G}_0)\sum_{t=1}^T \nabla_{\theta}\log p_{\theta}(\textbf{G}_{t-1}|\textbf{G}_t) \right] $$

We find that this formulation is inadequate and does not converge as the number of nodes increases, and we will include a plot in the paper showing reducing performance as the nodes increase.

Our key improvement is to modify the gradient in the following way:

$$ \nabla_{\theta}\mathcal{J}(\theta) = E_{\tau}\left[r(\textbf{G}_0)\sum_{t=1}^T \nabla_{\theta}\log p_{\theta}(\textbf{G}_{0}|\textbf{G}_t) \right] $$

Here, $\tau$ is the generation trajectory $\textbf{G}_{0:T}$.

By changing $\nabla_{\theta}\log p_{\theta}(\textbf{G}_{t-1}|\textbf{G}_t)$ to $\nabla_{\theta}\log p_{\theta}(\textbf{G}_{0}|\textbf{G}_t)$, we encourage the model to converge as close as possible to $\textbf{G}_0$, which has the highest reward at all timesteps. We find that skipping the complexity of considering all the transitions between the intermediate states $\textbf{G}_t \to \textbf{G}_{t+1}$ leads to a significantly reduced amount of fluctuations and better convergence.

The parameter is then updated using a learning rate $\eta = 10^{–5}$:

$$ \theta \gets \theta + \eta \cdot \nabla_{\theta}\mathcal{J}(\theta) $$

{\bl
\subsection{Hyperparamters}

For unconditional generation, we use the following hyperparameters: 

\begin{itemize}
\item $\kappa_1 = \kappa_2 = 0$
\item $\kappa_3 = 1$
\item $C = 0.7 r_{val} + 0.2 r_{nov}$ 
\end{itemize}

For targeted generation, we use the following hyperparameters:

\begin{itemize}
\item $\kappa_1 = 0.4$
\item $\kappa_2 = 0.3$
\item $\kappa_3 = 0.2$
\item $C = 0.7 r_{val} + 0.3 r_{nov}$
\end{itemize}

These are stated in reference to Eq (3) in the main text. 
}

{\bl
\section{Algorithm Tables}

\subsection{Training TransE on BioKG with SLCWA}

\begin{algorithm}
\label{alg:TransE_BioKG_SLCWA}
\begin{algorithmic}[1] 
\REQUIRE BioKG dataset $\mathcal{G} = \{(h, r, t)\}$, embedding size $d = 256$, learning rate $\alpha = 0.005$, margin $\gamma = 0.2 $, number of epochs $N=100$
\ENSURE Trained TransE embeddings for entities and relations

\STATE \textbf{Initialize}:
\STATE \quad Entity embeddings $E \in \mathbb{R}^{|E| \times d}$ randomly
\STATE \quad Relation embeddings $R \in \mathbb{R}^{|R| \times d}$ randomly

\STATE \textbf{Set Optimizer}: Adam with learning rate $\alpha$

\FOR{epoch $= 1$ to $N$}
    \FOR{each triple $(h, r, t) \in \mathcal{G}$}
        \STATE \textbf{// Generate negative samples using SLCWA}
        \STATE Select a corrupted triple $(h', r, t)$ or $(h, r, t')$ where $h'$ or $t'$ is sampled locally
        \STATE \textbf{// Compute scores}
        \STATE $score_{pos} \leftarrow \| E[h] + R[r] - E[t] \|_2$
        \STATE $score_{neg} \leftarrow \| E[h'] + R[r] - E[t'] \|_2$
        \STATE \textbf{// Compute margin-based loss}
        \STATE $loss \leftarrow \max(0, score_{pos} + \gamma - score_{neg})$
        \STATE \textbf{// Backpropagate and update embeddings}
        \STATE Compute gradients of $loss$ w.r.t $E[h]$, $E[t]$, $E[h']$, $R[r]$
        \STATE Update $E[h]$, $E[t]$, $E[h']$, $R[r]$ using Adam optimizer
    \ENDFOR
\ENDFOR

\end{algorithmic}
\end{algorithm}

\subsection{Score-Based Diffusion Model}

\begin{algorithm}
\label{alg:score_diffusion}
\begin{algorithmic}[1]
\REQUIRE Node feature matrix $ \mathbf{X} $, Adjacency matrix $ \mathbf{A} $, Number of diffusion steps $ T $, Score network parameters $ \theta $, Guidance scale $\lambda$, Context vector $\textbf{c}$
\ENSURE Trained score network parameters $ \theta $

\STATE \textbf{Initialize} score network parameters $ \theta $

\FOR{epoch = 1 to $ N_{\text{epochs}} $}
    \FOR{batch in data}
        \FOR{$ t = 1 $ to $ T $}
            \STATE Sample noise $ \epsilon \sim \mathcal{N}(0, \mathbf{I}) $
            \STATE Compute noisy features $ \mathbf{X}_t = \sqrt{\alpha_t} \mathbf{X} + \sqrt{1 - \alpha_t} \epsilon $
            \STATE Compute score: $ \mathbf{s}_\theta(\mathbf{X}_t, \mathbf{A}, t) = \nabla_{\textbf{G}_t}\log p_{\theta}(\textbf{X}_t, \textbf{A}_t) $
            \STATE Update guidance: $s'_{\theta} \leftarrow s_{\theta} + \lambda\nabla_{\textbf{G}_t}\log p_{\theta}(\textbf{c}|\textbf{X}_t, \textbf{A}_t)$
            \STATE Compute loss $ \mathcal{L} = \| \mathbf{s}'_\theta(\mathbf{X}_t, \mathbf{A}, t) + \frac{\epsilon}{\sqrt{1 - \alpha_t}} \|^2 $
        \ENDFOR
        \STATE \textbf{Update} $ \theta $ using optimizer with gradient $ \nabla_\theta \mathcal{L} $
    \ENDFOR
\ENDFOR

\STATE \textbf{Return} $ \theta $
\end{algorithmic}
\end{algorithm}

}

\end{document}